\documentclass[sigconf]{acmart}
\AtBeginDocument{\providecommand\BibTeX{{\normalfont B\kern-0.5em{\scshape i\kern-0.25em b}\kern-0.8em\TeX}}}

\copyrightyear{2023} 
\acmYear{2023} 
\setcopyright{rightsretained} 
\acmConference[CHI '23]{Proceedings of the 2023 CHI Conference on Human Factors in Computing Systems}{April 23--28, 2023}{Hamburg, Germany}
\acmBooktitle{Proceedings of the 2023 CHI Conference on Human Factors in Computing Systems (CHI '23), April 23--28, 2023, Hamburg, Germany}\acmDOI{10.1145/3544548.3581463}
\acmISBN{978-1-4503-9421-5/23/04}

\acmConference[CHI2023]{CHI '23: ACM Symposium on Human Factors in Computing}{2023}{Hamburg, Germany}
\acmBooktitle{CHI '23: ACM Human Factors in Computing, 2023, Hamburg, Germany}
\acmPrice{15.00}
\acmISBN{978-1-4503-9421-5/23/04}

\usepackage{graphicx,verbatim,tabularx,setspace,xspace}
\usepackage{xcolor}
\usepackage{array}
\usepackage{boldline}
\usepackage{subcaption}
\usepackage{multirow}
\usepackage{pdflscape}
\usepackage{pifont}

\usepackage{rotating}
\usepackage[utf8]{inputenc}
\usepackage{geometry}
\usepackage{ragged2e}
\usepackage{booktabs, tabularx}
\usepackage{enumitem}
\usepackage{etoolbox}
\usepackage{makecell}

\usepackage{pgfplots}
\pgfplotsset{compat=newest}
\usetikzlibrary{patterns}

\usepackage{url}

\newcolumntype{L}[1]{>{\raggedright\arraybackslash}p{#1}}

\newcolumntype{C}[1]{>{\centering\arraybackslash}p{#1}}

\newcolumntype{R}[1]{>{\raggedleft\arraybackslash}p{#1}}

\usepackage{etoolbox} \newtoggle{comments} \settoggle{comments}{true}
\iftoggle{comments}{
\usepackage[normalem]{ulem} \newcommand{\notejs}[1]
    	{{\color{brown}[{\bf Jat:} #1]}}
    \newcommand{\noteml}[1]
    	{{\color{blue}[{\bf Michelle:} #1]}}
	\newcommand{\notejc}[1]
    	{{\color{purple}[{\bf Jennifer:} #1]}}
    
     \newcommand{\notecn}[1]
    	{{\color{green}[{\bf Chris:} #1]}}
    \newcommand{\todo}[1]
    	{{\color{red}[{\bf Todo:} #1]}}	
    	
    \newcommand{\cut}[1]
    	{{\color{red}\sout{#1}}}

}{
    \newcommand{\notejs}[1]{}
    \newcommand{\noteml}[1]{}
    \newcommand{\notejc}[1]{}
    \newcommand{\notecn}[1]{}
    \newcommand{\todo}[1]{}
    \newcommand{\cut}[1]{}

}

\def\genbox#1#2#3#4#5#6{\leavevmode\raise#4bp\hbox to#5bp{\vrule height#5bp depth0bp width0bp
    \pdfliteral{q .5 w \csname #2COLOR\endcsname\space RG
                       \csname #3PDF\endcsname{#5}{#6} S Q
             \ifx1#1 q \csname #2COLOR\endcsname\space rg 
                       \csname #3PDF\endcsname{#5}{#6} f Q\fi}\hss}}

\def\sqbox      #1#2{\genbox{#1}{#2}  {sq}       {0}   {4.5}  {2.25}}
\def\trianbox   #1#2{\genbox{#1}{#2}  {trian}    {0}   {5}    {2.5}}

\begin{document}

\title{Out of Context: Investigating the Bias and Fairness Concerns of ``Artificial Intelligence as a Service''}

\author{Kornel Lewicki, Michelle Seng Ah Lee, Jennifer Cobbe, Jatinder Singh}
\affiliation{\institution{Compliant \& Accountable Systems Group}
\country{University of Cambridge}
}

\renewcommand{\shortauthors}{Lewicki et al.}

\begin{abstract}
``AI as a Service'' (AIaaS) is a rapidly growing market, offering various plug-and-play AI services and tools. AIaaS enables its customers (users)---who may lack the expertise, data, and\slash or resources to develop their own systems---to easily build and integrate AI capabilities into their applications. Yet, it is known that AI systems can encapsulate biases and inequalities that can have societal impact. This paper argues that the context-sensitive nature of fairness is often incompatible with AIaaS' ``one-size-fits-all'' approach, leading to issues and tensions. Specifically, we review and systematise the AIaaS space by proposing a taxonomy of AI services based on the levels of autonomy afforded to the user. We then critically examine the different categories of AIaaS, outlining how these services can lead to biases or be otherwise harmful in the context of end-user applications. In doing so, we seek to draw research attention to the challenges of this emerging area.

\end{abstract}



\maketitle

\section{Introduction}\label{Introduction}

Artificial Intelligence is increasingly being offered ``as a Service'' (AIaaS). 
Marketed as ``AI with no machine learning (ML) skills required'' \cite{AmazonAIServices}, AIaaS offers its users---primarily business customers---access to state-of-the-art AI capabilities, without the need for volumes of training data, expensive computational resources or lengthy development timescales that are generally required by traditional `in-house' machine learning development.

At the same time, the reach and extent to which AI algorithms now pervade our daily lives has put AI under increased scrutiny. Various high-profile controversies (e.g. \cite{buolamwini2018gender, ProPublica, obermeyer2019dissecting}) have raised concerns over the technology’s potential to both perpetuate existing societal inequalities and introduce new types of discrimination \cite{o2016weapons, garcia2016racist, howard2018ugly}.
As such, there are growing demands for greater levels of fairness, transparency and accountability over AI powered technologies (e.g., see \cite{dwivedi2021artificial, wieringa2020account, mitchell2019model, cobbe2021artificial}).

In this context, the emergence of the AIaaS paradigm presents a number of challenges. First, given that these services aim to offer sophisticated AI capabilities at low cost and on demand to potentially anyone, often requiring little to no technical expertise, there is a risk that any algorithmic biases or other undesirable behaviours in the AIaaS system could be reproduced on \textit{a far greater scale} across organisations \cite{cobbe2021artificial, javadi2021monitoring}. Further, with AIaaS products developed and sold by private for-profit companies, the inner workings of these commercial algorithms are often opaque or hidden (for various reasons), making it difficult for potential users to determine \textit{if}, \textit{when} and \textit{why} bias manifests itself within these services. 

Moreover, with an aim to appeal to the widest possible group of customers, many of these AI services are offered as generic `{AI building blocks}', aiming to underpin a wide range of customers' applications in a variety of different contexts. However, as existing literature makes clear, \textit{fairness is often contextual} \cite{selbst2019fairness, green2018myth, lee2021landscape, sambasivan2021re}:
a seemingly ``fair'' model designed for one context might be \textit{misleading, inaccurate, or harmful} when applied to a different context \cite{selbst2019fairness}. This points to an inherent tension between the nuanced context-sensitive nature of fairness and the generic, `one-size-fits-all' principle underpinning AIaaS.

In line with this, in this work, we treat the AIaaS paradigm as the object of study, considering whether enough thought has been put into the potential fairness risks associated with the development and adoption of AIaaS by an ever larger number of organisations. In particular, we question \textbf{whether the generic nature of such services \textit{is} and \textit{can be}  appropriate} given the diversity (i) of the potential applications in which they might be employed, and (ii) of the people, contexts, values, views and beliefs inherent to the real world. 

Towards this, we first propose a taxonomy of AIaaS offerings based on three decreasing levels of autonomy and control afforded to the user: (1) \textit{AutoML Platforms}, (2) \textit{AI APIs}, and (3) \textit{Fully-managed AI Services}. Through a combination of experimental evaluation, a consideration of topical examples,  and theoretical analysis of a subset of AIaaS services from leading providers, we then critically examine each of the different categories of AIaaS offerings, outlining the novel fairness concerns and practical challenges arising from the development and use of these services in practice. Specifically, we consider the \textit{perspective of the users} (customers) of these services (who can vary depending on the service type) and explore how the assumptions and constraints of different AIaaS models---which vary in levels of abstraction, transparency and control afforded by the provider to the user---can come into conflict with the contextual (and domain-specific) nature of fairness. This raises the risk of bias being embedded and propagating through the socio-technical systems supported by these services. Finally, we conclude by reflecting on what these concerns mean for the design, use and governance of AIaaS, identifying open challenges and suggesting potential paths forward.

In all, this paper (i) introduces and provides a taxonomy for describing various types of AIaaS services broken down by user perspectives so as to support an exploration of the concerns they raise; (ii) through the use of exemplars, elaborates the bias- and fairness-related issues and limitations of the {`one-size-fits-all'} approach of AIaaS given the diverse range of contexts in which such services may be used; and (iii) discusses the wider impacts of the popularisation of AIaaS models and highlights multi-disciplinary research opportunities in this space. Our broader aim is to draw community attention to this nascent yet increasingly influential area and encourage discussions and ways forward regarding  pragmatic approaches to the fair and responsible development and use of AI services. \section{Background}

To provide context for our exploration of fairness issues in AI services, we begin by briefly introducing AIaaS before describing some of the relevant emerging regulatory context, which both motivates the need for attention to be brought to issues of bias, {and indicates}
some of the challenges that AIaaS brings to the space. We then highlight, with reference to related work,  issues of algorithmic fairness and identify 
particular characteristics of AIaaS that can pose fairness-related challenges. 

\subsection{AI as a Service (AIaaS)}\label{Background:AIaaS}

There is much interest in the adoption of AI among both business and governments alike \cite{zhang2021ai, McKinsey2020StateOfAI, dwivedi2019artificial}. However, the nature of AI development---requiring volumes of training data, specialised hardware for building and training models, and technical expertise in machine learning---presents a significant challenge for organisations wishing to integrate AI into their business processes \cite{dwivedi2019artificial, cubric2020drivers}.

Recognising the opportunity, providers increasingly offer 
Artificial Intelligence solutions ``as a service'' (AIaaS) \cite{cobbe2021artificial, javadi2021monitoring}. AIaaS is advertised as a way of enabling developers without machine learning expertise to easily add AI capabilities to their applications. Various kinds of AI service exist (as we elaborate in~\S\ref{AIaaS}), providing different functionality (e.g. video and image analysis, speech recognition and synthesis, etc.), with differing levels of customisation, and likely attracting different users (from experienced data scientists looking to accelerate their workflows to business users with no prior ML-knowledge).
AIaaS promises its customers (the `\textbf{users}' of these services)\footnote{In this paper we use the term `users' to refer to customers of AI services.} access to ML solutions without the need for volumes of training data, expensive computational resources or lengthy development times that are required by the traditional machine learning project development process. Indeed, the dominant AI service providers tend to be those very companies with significant access to the data, infrastructure and expertise required for effective ML development, and therefore are those best placed to develop and provide industry-leading AI services to derive (additional) revenue. On the other hand, the users of the services---who may lack the required resources, data, time or technical know-how to undertake and develop their own AI---gain direct access to AI-driven capabilities that may otherwise be beyond their reach \cite{cobbe2021artificial}. 

Accordingly, and as was the case for the uptake of `traditional' cloud, it is likely that AIaaS will become \textit{the} primary means of AI proliferation and implementation across many industries and organisations (in contrast with ``in-house'' AI development)~\cite{MarketResearch, cobbe2021artificial}, given that such services reduce overheads and barriers to entry through providers offering and making accessible state-of-the-art tools\slash models. Indeed, we already see a consolidation of the AIaaS market predominantly around the few already dominant `Big Tech' companies (primarily from two regions: U.S. and China), which further entrenches their position and extends their societal influence \cite{cobbe2021artificial, ahmed2020democratization}. This places AIaaS providers---primarily the cloud technology giants---in a powerful position. Crucially, AIaaS moves providers beyond merely offering supporting infrastructure for applications (as in cloud services) to directly enabling, facilitating, and underpinning customers’ core applications and decision-making systems \cite{cobbe2021artificial}. As such, there is a risk that through their services, AIaaS providers might propagate not only unintended bias, but also their own set of views, values and priorities. This can have wide-ranging social consequences given the vast numbers of users of such services and because of the potential issues that can arise from the diverse range of contexts in which these ``generic'' services can be used. 

\subsection{AI Regulation}\label{Background:Regulation}
With AI adoption growing rapidly in recent years, there has only recently been specific regulation targeting AI systems (though the use of AI is subject to general legal frameworks, such as data protection~\cite{cobbe2021artificial} and non-discrimination~\cite{adams2022}). There is, however, growing consensus that  
AI-specific regulation is needed to help address the undesirable and potentially harmful effects and ramifications of AI technologies. This often builds on non-legal sets of principles or ethical standards proposed by academics, civil society groups, and others for `\textit{responsible}' and `\textit{fair}' AI \cite{dafoe2018ai, jobin2019global, schiff2020s, OECDLegal, MSFTApproach}. 

In the US, for instance, a proposed Algorithmic Accountability Act would require impact assessments for AI systems---including in relation to fairness and bias concerns around automated decision-making systems---and seeks to address some transparency and accountability issues around AI~\cite{aaa2022, mokander2022}. In the EU, a broader regulatory framework has been proposed in the form of the Artificial Intelligence Act (AI Act). Under the EU's proposed framework~\cite{EURLex5241} systems which are deemed to pose an \textit{unacceptable} risk (such as certain kinds of social scoring by public bodies) would be prohibited, while those responsible for systems classed as posing a \textit{high} risk would be subject to risk management, transparency, accountability, and other requirements, responsibilities, and obligations. These include specific obligations to address concerns around fairness and bias, targeting both the datasets used to train models and the ongoing operation of systems. Those responsible for lower risk systems face more limited responsibilities, while systems in the lowest category of risk are effectively exempt

While the increased legal and regulatory focus on fairness and bias concerns is welcome, neither the proposed Algorithmic Accountability Act in the US nor the EU's proposed AI Act properly account for the data-driven `supply chain' dimension that AI services bring. The Algorithmic Accountability Act would apply to ``covered entities'' who ``deploy'' AI systems for ``critical'' decision-making (provided they meet certain other criteria, such as around revenue and size of customer base) and \textit{also} to those who make systems for that purpose~\cite{aaa2022}. The AI Act will apply most of its requirements to ``providers'' of AI systems, who may in some cases be the users of AIaaS providers~\cite{EURLex5241} (we later explore the fairness implications of this (\S6.3)). Yet the divisions of responsibilities envisaged in these proposals do not address the fact that one provider will have many customers (users), each with their own contexts, use cases, and set of risks specific to their application. 
Moreover, these proposed laws do not address the information, skills, and power imbalances often present between AIaaS providers (again, typically large technology companies) and their service users (who may be much smaller organisations and\slash or  lacking certain degrees of technical expertise), which may make it difficult if not impossible in practice for AIaaS users to identify and mitigate problems with systems that are not under their control. Nor do they address the more fundamental fact that such services are not under their control in the first place. 

\subsection{Algorithmic Fairness}\label{Background:Fairness}

As AI becomes increasingly pervasive, its propensity to both amplify existing biases and social inequities and create new ones has attracted considerable attention across a range of communities, including academics, policy-makers, industry, and civil society. Much of the initial work focused on developing quantitative definitions of fairness (see, e.g., \cite{dwork2012fairness, liu2017calibrated, joseph2016fairness, hardt2016equality, verma2018fairness}), and various technical methods for `{debiasing}' AI models according to these mathematical formalisations (see, e.g., \cite{feldman2015certifying, bolukbasi2016man, agarwal2018reductions, zafar2017fairness}). \textit{De-biasing} refers to the practice of removing undesired skews in the data and the model outcome, such as by equalising a metric of interest between groups. In academic literature, ``unintended bias'' is used to describe the different \textit{sources} of harm that are introduced throughout an AI development lifecycle \cite{suresh2019framework, lee2021algorithmic}. A related taxonomy is the types of `harm' as the consequences and scale of impact of AI \cite{crawford2016artificial}. While the former focuses on \textit{how} the harm was introduced in the lifecycle, such as through poor data collection mechanisms, the latter describes \textit{what type} of harm resulted from the unintended biases. While these taxonomies generalise on types of harms and their sources, the nuances of whether an algorithm is harmful and whether the potential risks of harm outweigh the potential benefits depend on the use case. More recently, aided by the interdisciplinary research in this area, there has been a growing realisation that \textit{fairness is often contextual}; where considerations can differ across application domains, regions, cultures, and so forth,  while some harmful algorithmic behaviours may only arise, or may only be recognised as harmful, when a system is used in a particular way, or in the presence of particular social or cultural dynamics \cite{selbst2019fairness, green2018myth, lee2021landscape, sambasivan2021re}. Accordingly, given the many complexities of fairness and its contextual nature, it is generally not possible to fully debias an AI system or to guarantee its fairness \cite{mehrabi2019survey, pleiss2017fairness, kleinberg2016inherent}; rather, the aim is to \textit{mitigate} fairness-related harms and other unwanted consequences as much as possible \cite{mehrabi2019survey, sun2019mitigating, selbst2019fairness}.

A growing body of interdisciplinary work studies ML fairness through a sociotechnical lens, which is aware of and actively considers the multitudes of social perspectives, actors, and interactions involved \cite{selbst2019fairness}. In the Human-Computer Interaction (HCI) domain, there has been work on studying public perceptions and expectations related to fairness in algorithmic systems. For instance, Binns et al.~\cite{binns2018s} and Woodruff et al.~\cite{woodruff2018qualitative} identify gaps and dissonances in public understanding of this concept, while Srivastava et al.~\cite{srivastava2019mathematical} found that these do not always align with existing mathematical definitions for fairness. Other work has focused on organisational challenges and barriers that practitioners face when attempting to build more responsible AI products and services \cite{rakova2021responsible, madaio2020co, madaio2022assessing}, as well as considerations regarding fairness perceptions across cultures \cite{sambasivan2021re}. 
Research has also been directed specifically towards better understanding AI practitioners' needs and the development of frameworks, processes and tools to help assess and audit algorithmic systems for unfair, biased, or otherwise harmful behaviour (e.g., \cite{raji2020closing, metaxa2021auditing, bird2020fairlearn, bellamy2019ai, madaio2020co}) -- both internally (within the organisations responsible for developing and maintaining these systems) and externally (by independent auditors, users and/or regulators). Particularly relevant here is the work exploring the issues and efficacy of fairness-specific tooling for supporting practitioners, for example, that of Holstein et al. \cite{holstein2019improving}, Lee et al. \cite{lee2021landscape} and Deng et al. \cite{deng2022exploring}, which considered the perceptions and use of so-called ``fairness toolkits'' that aim to support ML practitioners with fairness concerns, finding significant disconnects between the tooling and the expectations, needs and practices of practitioners.

\subsubsection{Fairness and AIaaS}\label{Background:Fairness:AIaaS}

Though issues of fairness pervade AI in general, there is the potential for AIaaS---in making widely-accessible, scalable and readily-available AI that is capable of underpinning a broad range of applications---to amplify AI fairness problems and introduce additional risks. Like any AI system, AI services can compound existing inequities by producing unfair outcomes, reinforcing pernicious stereotypes and disproportionately distributing negative consequences of technology to those already marginalised. 

Each provider’s AI services are offered to a range of possible customers, each of which can use the services for widely varying purposes. As such, issues of bias can arise both from the nature of the underlying ML model itself \textit{and} the operating context in which a user employs that service. This means that  AIaaS 
might not only contribute to {replicating} fairness issues at scale by propagating model's intrinsic biases (or other issues) to the vast number of user applications powered by a given service (see  previous work including \cite{buolamwini2018gender, scheuerman2019computers, koenecke2020racial,singh2019decprov}). Rather, it also raises challenges in terms of both identifying and mitigating fairness issues that can arise from a user's particular application of the service. This is because it can be difficult (if not impossible) to preempt all possible purposes for which a service might be used, while users employing these services may not have the knowledge, expertise, ability or access to assess the appropriateness of the service within their context and employ relevant safeguards. That is, as we discussed previously, one of the key difficulties in uncovering and mitigating unintended bias in AI systems is that many fairness issues will manifest only in particular contexts, meaning that understanding, let alone accounting for all bias concerns in an effort to create a `generally applicable' AI service appears infeasible.

Importantly, much of the relevant literature on fairness, bias, and related harms has focused on AI systems built `\textit{in-house}', where it is generally assumed that those producing the ML system are actively  involved in building the model, and can adjust the model and/or the data it is trained on to mitigate fairness-related harms. There is some work on fairness relating to AI services, though this tends to focus on either exposing or `auditing' certain AI services \cite{buolamwini2018gender, raji2019actionable, scheuerman2019computers, de2019does}, or devising tools, methods and interventions relevant for those building (rather than \textit{using}) such services \cite{mitchell2019model, arnold2019factsheets, 
raji2020closing, madaio2022assessing}. 
However, there has been rather less focus on fairness issues and concerns in AI services as regards the perspective of those using the models and services of others.
In arguing this as an area warranting more attention, in this paper we explore fairness from the perspectives of the users (customers) of AIaaS and highlight potential issues and challenges that arise from such services. 
 \section{AIaaS Taxonomy}\label{AIaaS}

  \begin{table*}[ht]
    \footnotesize
    \centering
    \begin{tabularx}{\linewidth}{@{} >{\RaggedRight\hsize=0.3\hsize}X
                                *{3}{>{\RaggedRight\hsize=1.1\hsize}X} @{}}
        \toprule
        & \textbf{AutoML Platforms}   & \textbf{AI APIs}  & \textbf{Fully-managed AI Services}    \\
        \midrule
    Description  &
        Automated model creation tools for building and deploying custom ML models.  & Prebuilt AI models and services available via pay-per-use APIs.            & AI models, services and platforms created and managed by an external third-party.    \\ \addlinespace
    \midrule
    Use Cases &
      Predominantly tabular data classification and regression tasks. &  Predominantly unstructured data problems such as object recognition, machine translation, text analytics etc.  & Industry-specific problems such as candidate screening, recidivism prediction,  healthcare allocation etc.  \\ \addlinespace
    \midrule
    Control &
      User maintains control over the data used to train the model and some training configuration, while remaining  steps of model creation are automated. &  Models are already pretrained by the provider and available for use. The user has only limited or no control over the model.  & Virtually entire control over the service is delegated to the service provider and the user interacts with the service via vendor provided ``no code" UI.  \\ \addlinespace
    \midrule
    Providers &
      Largest cloud providers and \newline dedicated AutoML startups. &  Predominantly largest cloud providers.  & Dedicated tech companies and startups.  \\ 
      \midrule
    Level of technical expertise  &
    
      Data scientists and developers with at least \textit{some} AI knowledge. &  Developers with technical but not necessarily AI knowledge.  & Minimal; suitable for those with non-technical backgrounds (e.g. general business users). \\
      \bottomrule
    \end{tabularx}
    \caption{AIaaS Taxonomy including three types of AIaaS services and their characteristics.}
    \label{table:comparisonAIaaS}
    \vspace{-4mm}\end{table*} 
To better understand and explore the fairness risks and challenges of ``\textit{AI as a Service},''  it is important to first consider the overloaded nature of the term. We therefore now introduce a taxonomy to help characterise the various types of AI services currently on offer, in terms of their functionality and how they represent the different degrees of automation, complexity and user involvement. The taxonomy serves to provide a more precise description of the offerings and to better enable the exploration of the potential fairness risks and harms of different types of AIaaS. 

Given AIaaS is an established business\slash marketing term,  our guiding principle in structuring this space and deriving the taxonomy was to understand and reflect the ways in which the AIaaS providers distinguish these services themselves. Accordingly, as a first step, we have compiled a list of relevant AIaaS providers by consulting prior research \cite{javadi2020monitoring, cobbe2021artificial, scheuerman2019computers, davenport2019potential}, market reports \cite{MarketResearch}, blogs \cite{yao_2022, kdnuggets} and news articles \cite{haranas_2022, zdnet}. Next, we comparatively reviewed the websites and marketing materials of AIaaS providers, elaborating the key characteristics, differences and similarities between the services offered, their naming, and the degree of user interaction and control. This process involved aligning the categorisation with the descriptions and terminology used in other AIaaS work \cite{javadi2020monitoring, javadi2021monitoring, xin2021whither, cobbe2021artificial}.

Table 1 represents the resulting taxonomy comprising three service  categories: (1) \textit{AutoML Platforms}, (2) \textit{AI APIs}, and (3) \textit{Fully-Managed AI services} (which we describe below). These categories reflect the perspective of users' (decreasing) level of involvement, required technical expertise and control over the service and the underlying ML model(s), as well as the increasingly specialised types of problems they address. These categories are not mutually exclusive in that some services may have characteristics that overlap the service types; however, identifying the service types helps indicate and highlight certain properties and characteristics that can have fairness implications in practice.

Consistent with other, more traditional cloud ``\textit{as a Service}'' offerings (such as \textit{SaaS} and \textit{PaaS}), we focus AIaaS as referring to a set of \textit{services} that are offered on demand, on a `for a fee' basis and target (primarily) business customers. Here, we do not consider open-source and other tools that can be downloaded, used and operated directly (i.e. in a \textit{non-service} context), such as AutoML frameworks (e.g. AutoKeras \cite{jin2019auto}, TPOT \cite{olson2016tpot}) and AI model libraries (e.g. Hugging Face \cite{HuggingFace}, ModelZoo \cite{ModelZoo}), nor Large Language Models (e.g. BERT \cite{devlin2018bert}, GPT-3 \cite{brown2020language}) or Generative AI tools (e.g. ChatGPT \cite{ChatGPT}, DALL-E \cite{DALLE}, Stable Diffusion \cite{StableDiffusion} etc.), though we recognise that some of the points we discuss may also be applicable to these and other forms of AI tooling. Rather, we focus  our  discussion on offerings provided as commercial services on an on-demand basis, which raise considerations and implications relating to expectations around (i) levels of user expertise; (ii) the degree of visibility, control and access a user has;  (iii) the potential for services to change and adapt (potentially without user knowledge); and (iv) the transactional (`per-use') nature of the service; and the legal and responsibility implications that come with commercial, data-driven, run-time supply chains involving several actors (\S \ref{Background:Regulation}). 

Also, note that our aim here is not to form a comprehensive, future-proof, or definitive taxonomy of AIaaS, but rather represents a first attempt at categorising this evolving, market-driven space in a way that helps highlight the issues and brings the debate on fairness in AI to include AIaaS. As a living document, the taxonomy can be adapted and evolved as new considerations are brought forward and according to the particulars of the vendors and services, regulations, and so on. We now describe each of these categories in turn.

\subsection{AutoML Platforms}

\textit{Automated Machine Learning} (AutoML for short) refers to the idea of building ML models with limited human intervention, either by automating {certain stages} in the ML workflow \cite{lam2017one, khurana2018feature, malkomes2016bayesian, kotthoff2019auto} or by automating the {entire} ML development process \cite{AmazonSageMaker, CloudAutoML, AutoMLAzure, 4Paradigm}. Here we focus on the latter, using ``AutoML Platforms'' services providing access to end-to-end, off-the-shelf AutoML functionality.
In this way, by automating virtually every step of ML pipeline construction, AutoML Platforms allow users to create custom ML models (potentially) without ever seeing a single line of code used to create those models. From the user involvement perspective, the AutoML Platform model creation process is limited to broadly three steps: (1) provision of the dataset to be used for the model training, (2) selection of the feature to be predicted, and (3) specification of the training budget {(the maximum time the AutoML service will spend training user's model)}. Likewise, the subsequent deployment of these models is often also limited to a simple click of a button, and the whole process is managed through a no-code UI.

In terms of providers, the current landscape of commercial AutoML Platforms is fast-growing and diverse. The commercial AutoML offerings can be categorised broadly into two groups: those offered by major cloud providers (e.g., Google, Microsoft), and  those offered by dedicated AutoML organisations (often startups, e.g., DataRobot, H2O.ai, 4Paradigm) \cite{xin2021whither}. The former exists as part of larger ecosystems of the cloud providers' broader offerings and are tightly integrated within them, while the latter, being more standalone, provide end-to-end support through their own offerings or by facilitating integration with external tools and platforms \cite{xin2021whither}.

\subsection{AI APIs}

AI APIs refer to a set of services offering access to a range of \textit{pre-built} ML models that users can essentially `plug' into their applications via pay-per-use third-party APIs. These services are offered on an ``as-is'' basis and as such are available for instant use. Generally, to use a particular model, a user sends to the AIaaS-provider a request (typically by a webservices API) specifying the desired service functionality (e.g. detect faces, categorise text), and the data to be treated as inputs (e.g. an image or text file). After conducting the required authentication and authorisation checks (API keys, budget, etc.), the provider will process the request and return the ML model's response \cite{javadi2020monitoring}. In this way, the AI APIs can be thought of as `AI building blocks', where the customer uses one or more of such services, integrating them into a  solution for their particular needs. 

With respect to the services offered,  we generally observe that the major AI APIs providers \cite{AmazonAIServices, GoogleCloudAI, IBMWatsonAI, MicrosoftCognitive} tend to focus on generic capabilities that are useful in a wide range of contexts, and broadly many offerings can be grouped into four key categories of service: 
\begin{itemize}
    \item \textit{Data Analytics} services - enable the analysis of users’s data and include capabilities for anomaly detection, content personalisation, product recommendation and so forth. 
    
    \item \textit{Text} services - offer AI capabilities in areas such as machine translation, text analytics, sentiment analysis, conversational interfaces etc.
    
    \item \textit{Speech} services - comprise  speech processing tools such as speech-to-text, text-to-speech, speech translation and speaker recognition.
    
    \item \textit{Vision} services - support content identification and analysis within image and video data. Example services include capabilities in object recognition, facial analysis, facial recognition, video indexing etc.
\end{itemize}

\subsection{Fully-managed AI Services}

Recently, we can also observe a growing trend of AI capabilities being offered in the form of fully-managed AI services. Whereas the aforementioned AI API services predominantly focus on `one-shot' functionalities (where a user sends over some input data and receives back the results of the ML model prediction for that particular data), here the offering tends to be more complex and focused, where the service seeks to supply a partial or complete business workflow. Fully-managed AI services might ingest a variety of a customer's resources and data, consist of several models and steps before producing a final prediction outcome, come with dedicated UI platforms or hardware, and even interact directly with the customer's own users. These offerings typically offer a more specialised set of functionalities targeting niche industry-specific needs, examples including algorithmic hiring \cite{HireVue, pymetrics}, medical processes such as diagnosis and triage \cite{QureAI, Lunit75:online, Infermedica}, and mental health self-care \cite{Headspace} to name but a few. Further, whereas the AI APIs generally need to be programmatically further integrated into users' own applications, fully managed services can also come as more complete products, providing dedicated mobile and web interfaces or desktop applications and involve multiple layers of processing. As such, the fully managed AI services model resembles that of a traditional SaaS (Software-as-a-Service) model, which removes the user from the low-level, technical complexities of these systems. \section{Methods}\label{Methods}

In this paper we elaborate the different AI service types, explaining some of their key characteristics from fairness perspective and some risks they entail. Our approach is based on a critical review of AIaaS vendors' claims and practices and empirical experimentation using select AIaaS services. 
For each service type, we outline a set of fairness risks and tensions that can occur as a result of (i) the decreasing levels of autonomy and transparency they afford to users, and at the same time (ii) increasing the level of  abstraction away from the social context in which these systems will be deployed.

In outlining these fairness concerns and considerations, we draw from the existing body of work on algorithmic fairness, and highlight the ways in which these issues are enacted and potentially amplified through AIaaS. We support our argument with a set of technical experiments using select AI services from leading providers to explore these issues in-depth and concretely demonstrate how they can manifest themselves in practice. Specifically, using three real-world datasets commonly used in the Fair-ML literature (Adult \cite{kohavi1996scaling}, German Credit \cite{king1995statlog} and COMPAS \cite{ProPublica}) we first explore how existing AutoML platforms---through their focus on optimising ML model performance according to a single, unconstrained, \textit{fairness-unaware} objective---can inadvertently lead to developing models with poor levels of fairness. 
Second, we explore issues of bias in the context of AI APIs and demonstrate why without knowing 
\textit{how} their services will be used, \textit{where, when} and by \textit{whom},
it is virtually impossible for these services to 
properly account for all possible concerns and issues that may be relevant.
We further perform {a} qualitative analysis of providers' claims and practices (e.g. what they have disclosed about the performance of their systems, development and validation procedures, and bias mitigation support), as has been done in specific algorithmic auditing contexts~\cite{raghavan2020mitigating}, to examine the various other trade-offs providers need to consider and highlight particular causes for concern. 

Importantly, the aim of our experiments is not to perform a comprehensive analysis of the bias issues in any particular AI service; and further, we stress that our exploration should be viewed as presenting simply a snapshot of current AIaaS landscape and some of its potential issues at a particular point in time (given these services may be updated or otherwise changed at any time).
Rather, we use practical, tangible exemplars to expose some fundamental issues concerning bias and fairness for the different AIaaS categories, which we argue restrict the utility and appropriateness of these services given the diversity of the values, views and beliefs inherent in the real world.  \section{{AIaaS} Fairness Concerns}

We now walk through and explore bias and fairness issues for the three AIaaS service types earlier described.

\subsection{AutoML Platforms}\label{Concerns:AutoML}

We first consider the AIaaS model offering the greatest levels of user autonomy, which are the AutoML platforms. In the AutoML setting, the goal of the AI service is to create \textit{custom} ML models in a fully-automated way. The user retains control over the training data used to build the model, but the remaining steps of the ML pipeline, from feature engineering to model selection and optimisation are automated (to varying degrees), with the user's involvement often limited to the selection of the feature to be predicted and the optimisation metric to be used when selecting the optimal model. On one hand, this approach allows users of all skill levels to quickly build state-of-the-art ML models; on the other hand, it also greatly limits their influence over model specifics. Further, {with} many commercial AutoML solutions tightly integrated within the service provider's infrastructure, user control over the created models is further restricted. We now discuss how these restrictions could cause tensions from the fairness perspective and constrain the ability for users to ensure fairness of the models built with such services.

\subsubsection{Fairness-unaware optimisation}\label{Concerns:AutoML:Optimisation}

\begin{figure*}
\begin{subfigure}{.33\textwidth}
  \centering
  \includegraphics[width=\linewidth]{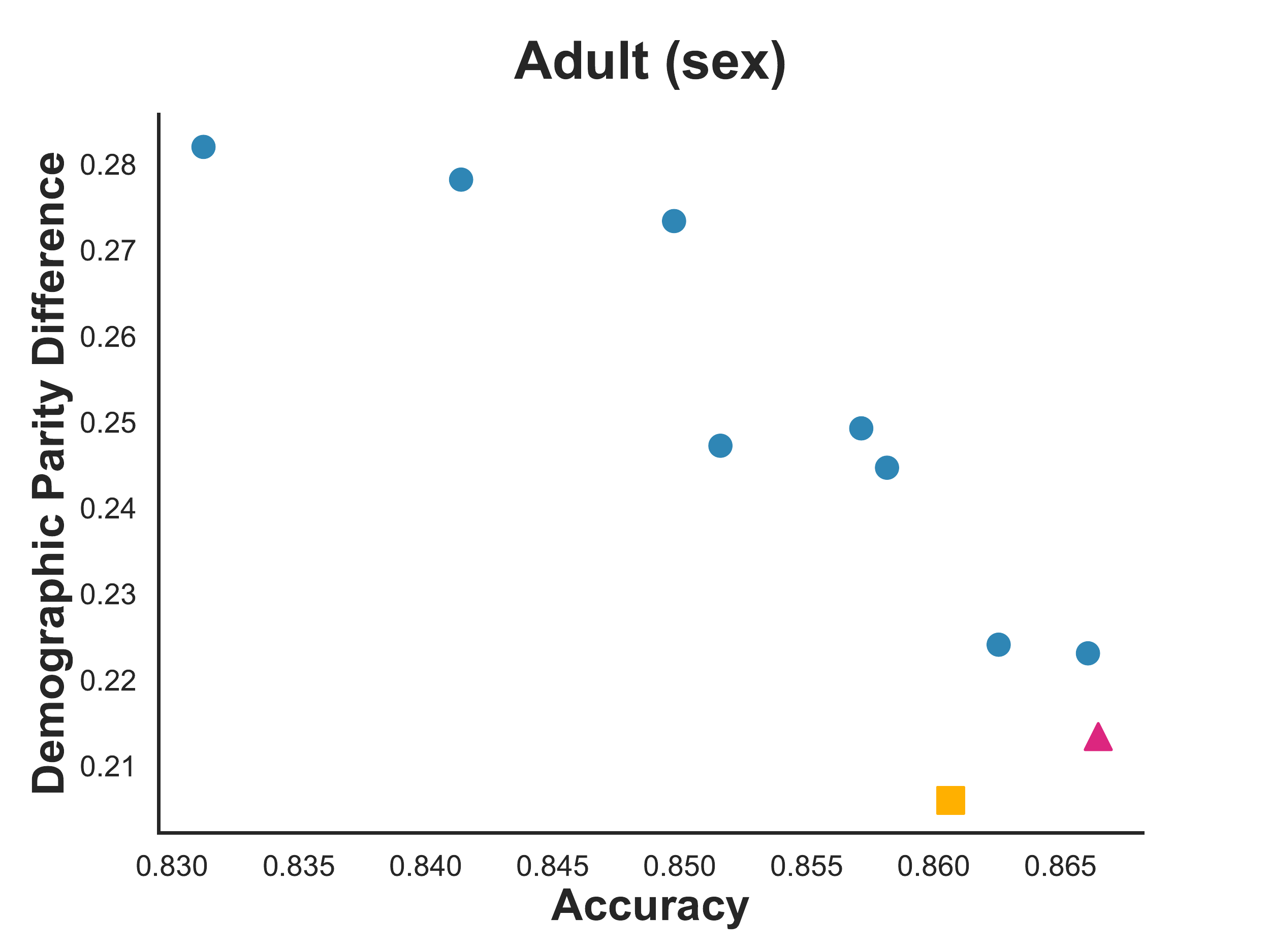}
\end{subfigure}\begin{subfigure}{.33\textwidth}
  \centering
  \includegraphics[width=\linewidth]{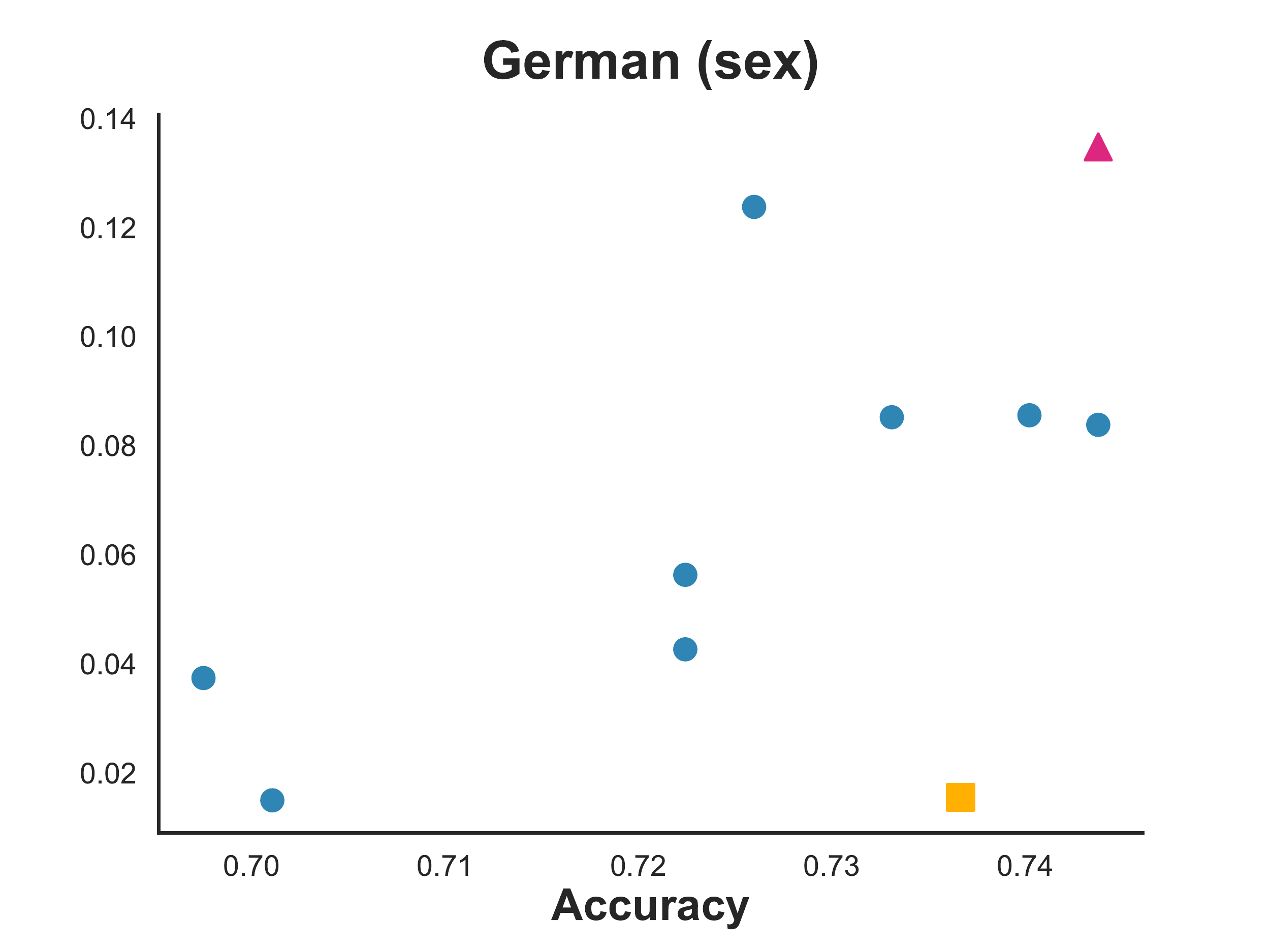}
\end{subfigure}
\begin{subfigure}{.33\textwidth}
  \centering
  \includegraphics[width=\linewidth]{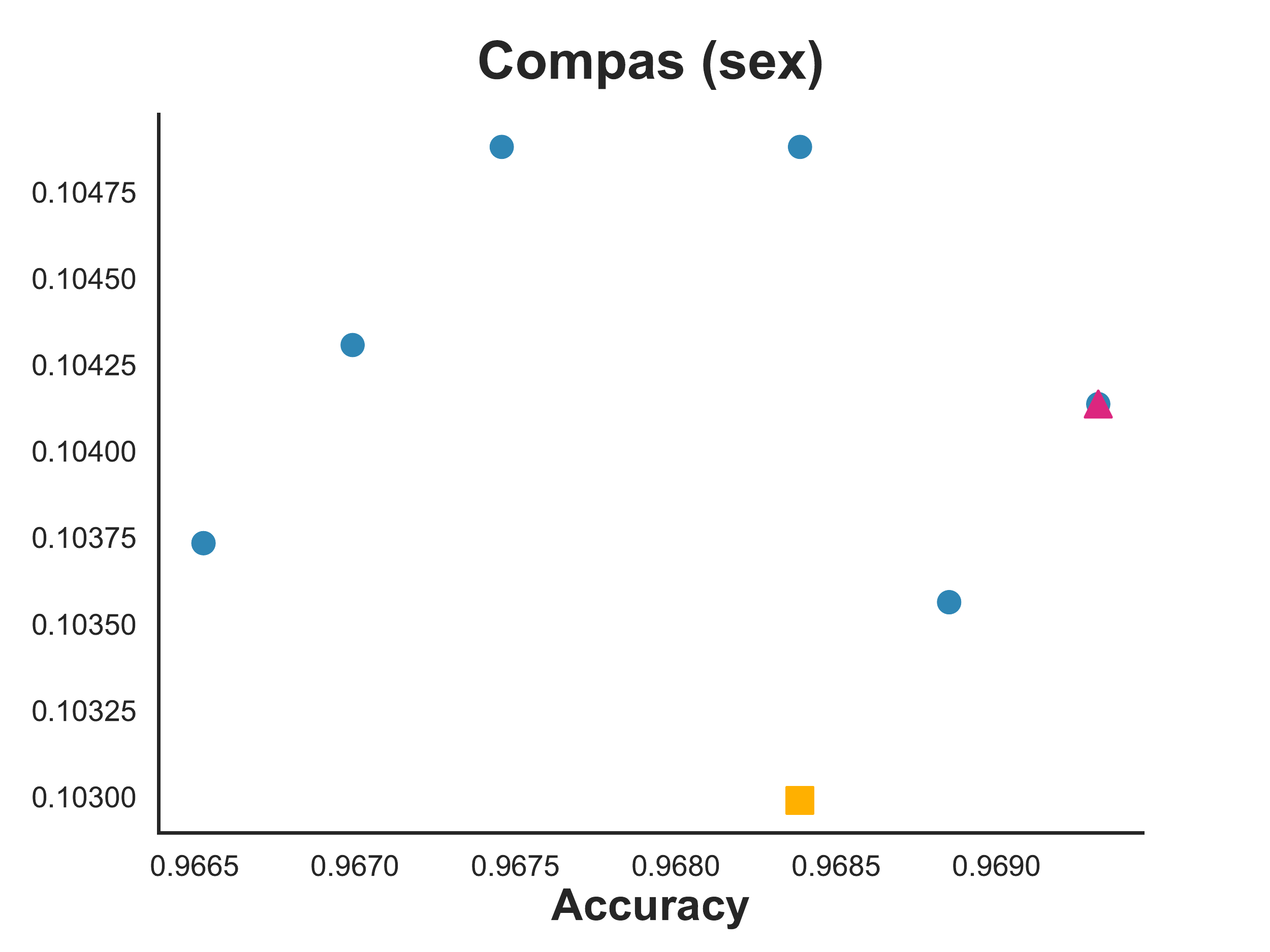}
\end{subfigure}
\begin{subfigure}{.33\textwidth}
  \centering
  \includegraphics[width=\linewidth]{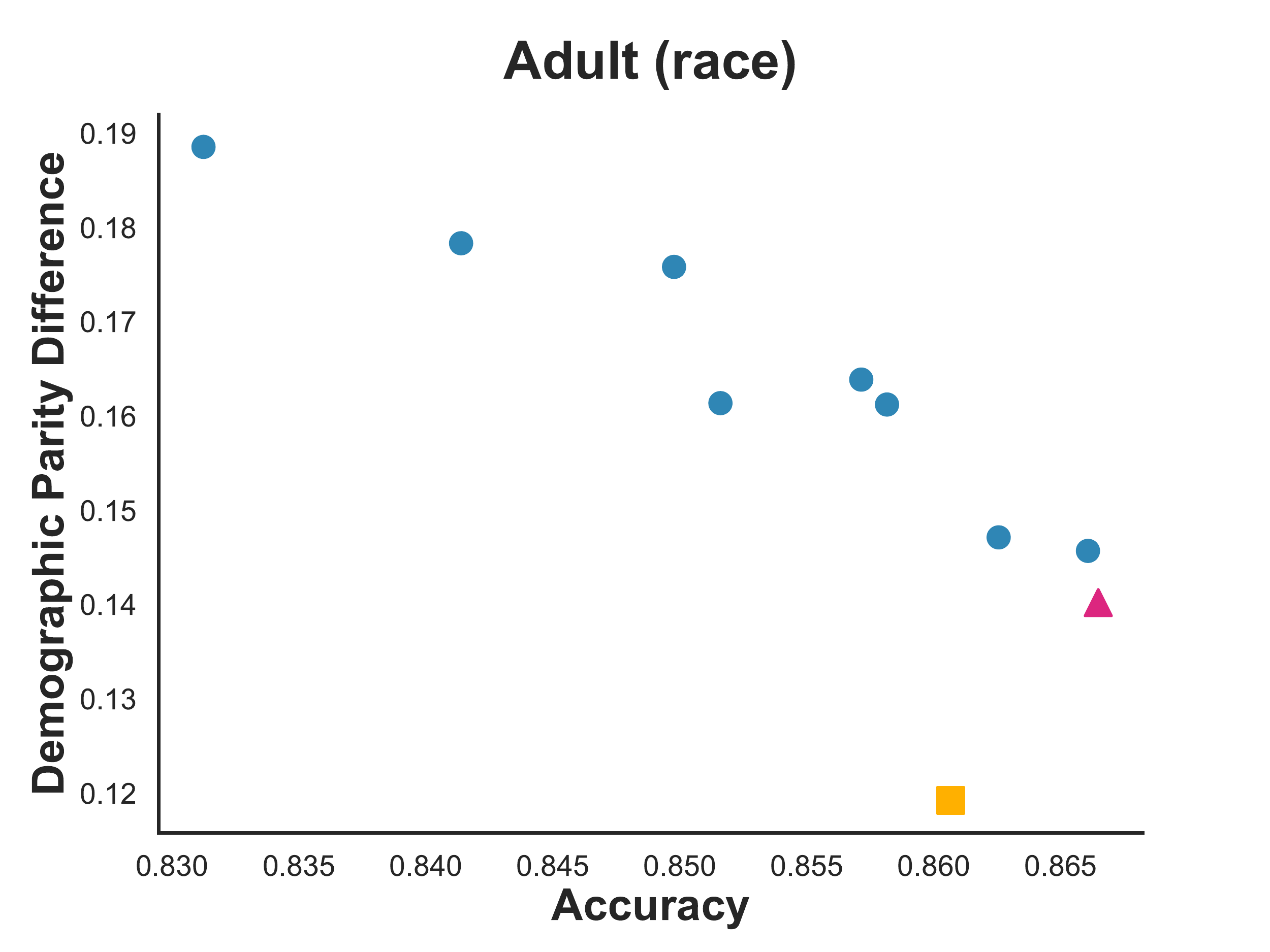}
\end{subfigure}\begin{subfigure}{.33\textwidth}
  \centering
  \includegraphics[width=\linewidth]{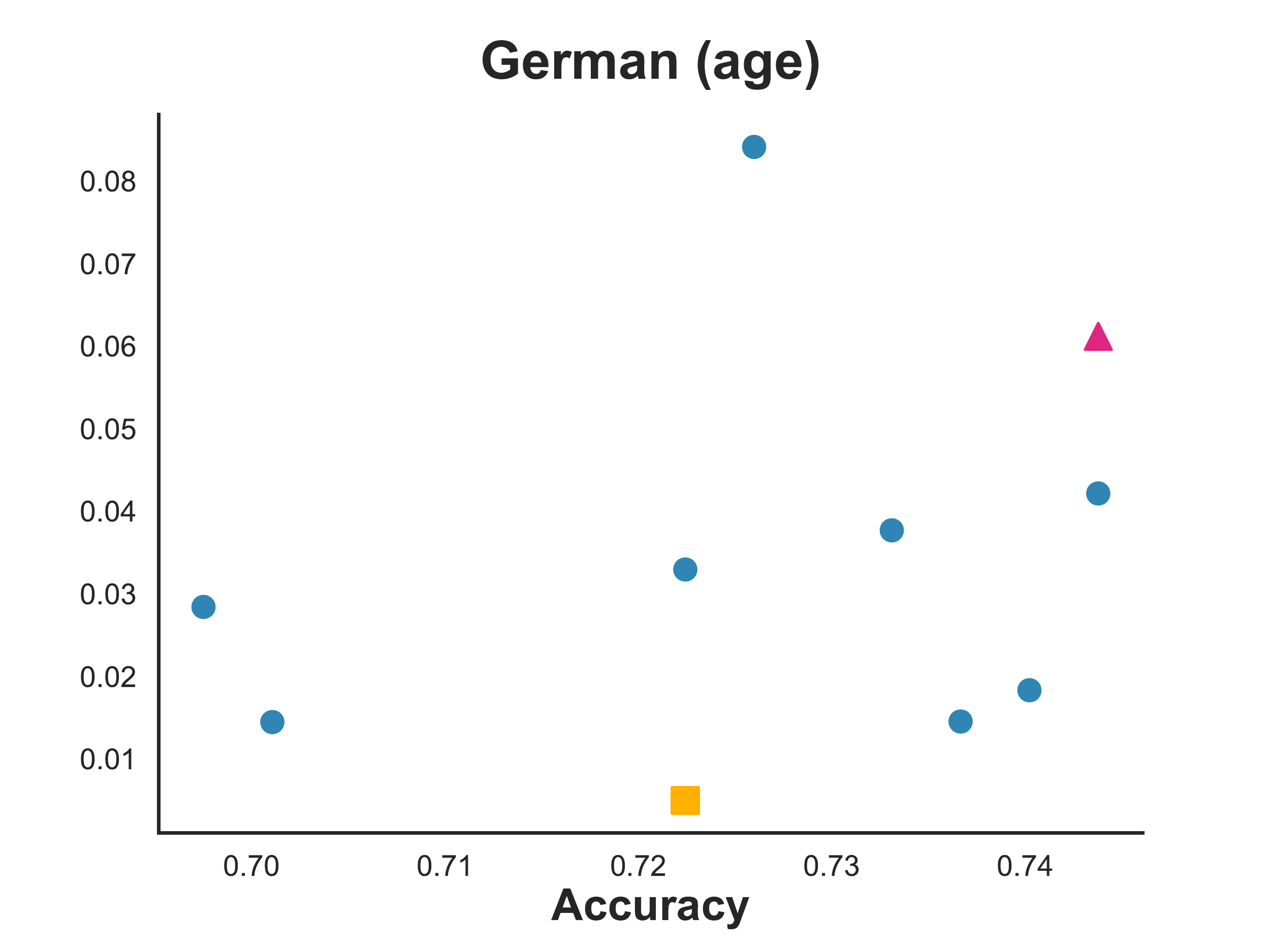}
\end{subfigure}
\begin{subfigure}{.33\textwidth}
  \centering
  \includegraphics[width=\linewidth]{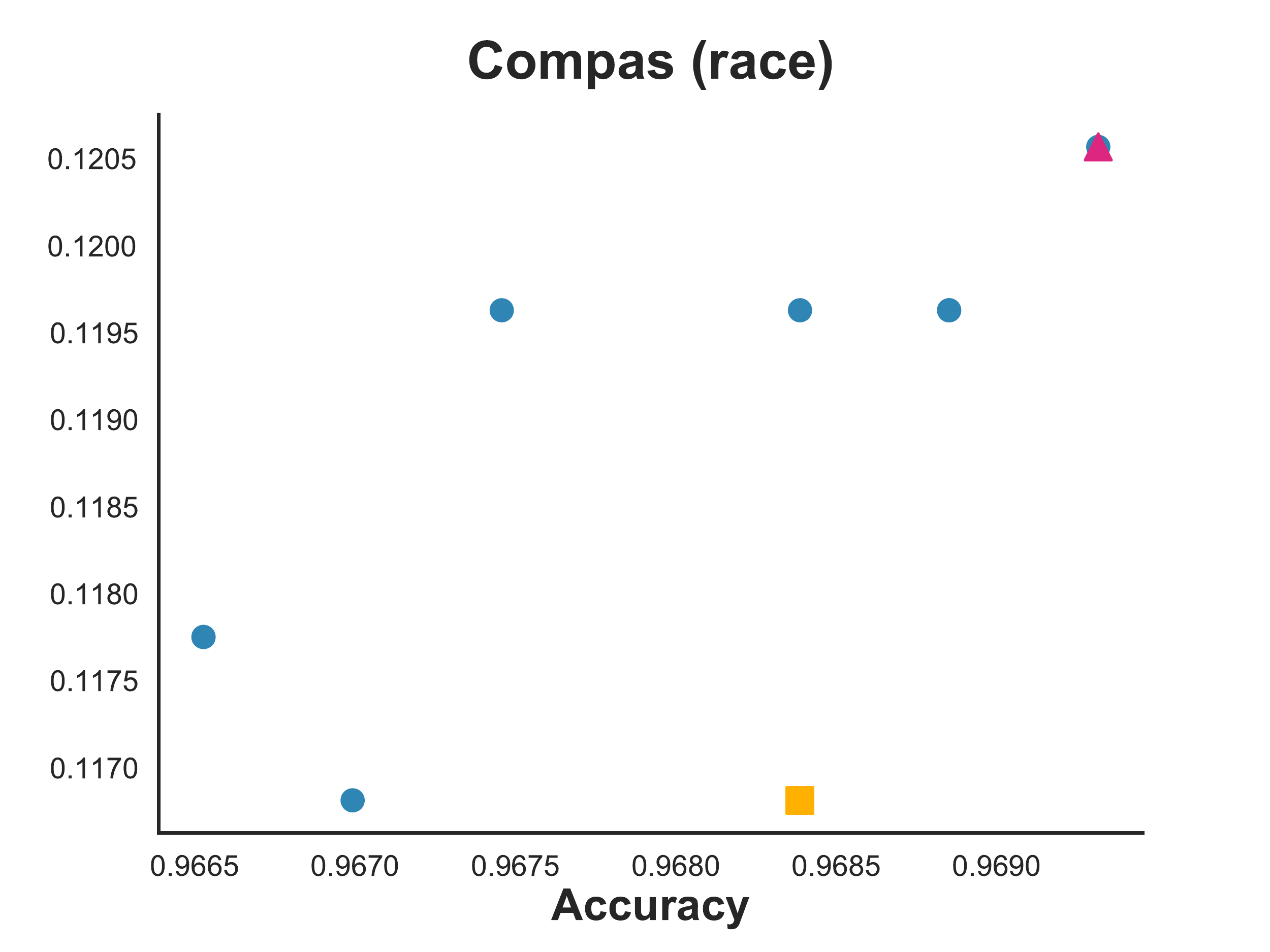}
\end{subfigure}

\caption[Fairness-accuracy trade-off of Azure AutoML models]{Fairness-accuracy trade-off of models created by Azure AutoML on Adult, German and COMPAS datasets measured across age, race and sex attributes. In pink (\trianbox1{cpink}), the model selected by the algorithm as ``optimal'' (higher is better); in orange (\sqbox1{corange}), the model which achieves the best fairness according to ``demographic parity difference'' metric (lower is better). As a result of selecting models solely by their predictive performance, the models deemed as ``optimal'' are often more biased than possible alternatives.}
\label{fig:fig}
\end{figure*} 
The fundamental premise of the AutoML paradigm is that the traditional `manual' model creation process can be replaced by an automated creation and subsequent search over possible model architectures in order to find the ``\textit{optimal}'' model. In the context of AutoML platforms, this `optimality' usually takes the shape of the best predictive performance as measured by one of the user selected metrics such as \textit{accuracy} or \textit{log loss} -- meaning that a model's fairness is typically not considered when choosing between the various models created in the AutoML process. However, for the many real-world applications where a model can impact people, it is not sufficient to only have high prediction accuracy. {Indeed, there is an increasing expectation and demand that ML practitioners ensure that the outcomes of ML applications are fair in that they do not discriminate against certain groups or individuals \cite{dwivedi2019artificial, cath2018governing}}. Unfortunately, as prior studies have shown \cite{cruz2020bandit, perrone2021fair, pfisterer2019multi, wu2021fair}, prioritising such fairness-unaware model optimisation---e.g. by focusing solely on the model's accuracy---can lead to inadvertently prioritising models with higher levels of bias.

To demonstrate the issues that `fairness-blind' optimisation can cause in the AutoML service context, we utilise Azure's AutoML service as an example service from a prominent provider (Microsoft), to build models for three real-world datasets commonly used in the fair-ML literature---Adult, German Credit and COMPAS \cite{kohavi1996scaling, king1995statlog, ProPublica}---and examine the resulting models' predictive accuracy and fairness. Figure 1 shows the fairness and performance of models created using the Azure AutoML service and highlights: (1) the model deemed by the platform as ``optimal'' and (2) the model achieving best fairness according to the \emph{demographic parity difference} metric \cite{dwork2012fairness}. 
We find that it is indeed often the case that as a result of selecting models solely by their predictive performance, the models deemed as ``optimal'' might in fact exhibit  more bias than possible alternatives (up to 12\% difference for German dataset split on the sex attribute), while `fairer' alternatives with only marginally lower levels of accuracy do exist.
This is particularly important when one considers that not all AutoML platforms offer customers the opportunity to {choose between} the various models that were created as part of the AutoML model creation and search process. For instance, in case of  Google's AutoML offering, the user is only provided with access to the final model, deemed by the service as ``optimal'', and does not have the opportunity to select other, potentially less-biased {(or otherwise more appropriate)} , models that were created during the training process, to use and deploy.

The current reliance of AutoML frameworks on fairness-blind optimisation metrics combined with the inability to choose between the various models created in the AutoML process abstract away the complex, domain-specific notion of `optimality' and its relational nature. In the face of the inevitable trade-offs between the real-world impact of changes in accuracy, fairness, privacy, consumer autonomy, and other considerations \cite{lee2021algorithmic}, such limitations restrict the users' ability to build a model that best reflects the values and objectives of  their particular use case.

\subsubsection{Constraints of proprietary platforms}\label{Concerns:AutoML:Constraints}

The constraints of the proprietary model of commercial AutoML platforms extend beyond the model training process to the model parameter tuning, model selection, deployment and monitoring. This affects the scope of actions users could take to further interrogate created models or adjust them should any issues be observed.

Consider bias mitigation. The fair-ML literature has introduced various techniques that change the source data (pre-processing) or the outputs (post-processing) \cite{feldman2015certifying, zhang2018mitigating, pleiss2017fairness}, as well as techniques that train a model to both maximise accuracy and increase fairness (in-processing)~\cite{martinez2020minimax, cruz2020bandit}, that could be used to mitigate potential bias issues we have outlined above. While, various open source ``fairness toolkits'' have been developed to make these fairness methods more widely accessible to model developers \cite{bird2020fairlearn, bellamy2019ai}, currently these are largely unsupported by most AutoML platforms. Instead, AutoML customers' choice is generally limited to the provider's own, proprietary, bias mitigation functionality. 

Yet again, in offering this functionality, AutoML providers make certain assumptions about how their services will be used and by whom, which may come into conflict with the customer's particular needs. For instance, IBM's AutoAI platform provides dedicated bias assessment and mitigation functionality in the form of its Watson OpenScale service allowing customers to uncover and mitigate bias according to the disparate impact fairness metric \cite{WatsonOpenScale}. However, no further fairness metrics are supported. Here, using IBM's service, the user would have no choice in defining the fairness metric that is suitable for their particular use case, and is instead forced to use the sole fairness definition defined by the platform, which may be inappropriate for their application domain.

Note, however, that even if AutoML providers decided to make or extend the set of various fairness metrics and mitigation techniques available, these optimise for just  narrow and particular definitions of fairness. Effectively tackling the fairness issues brought by real-world use cases requires more; indeed, addressing issues of fairness  is a contextually specific exercise, requiring careful consideration of the domain specifics \cite{lee2021landscape, selbst2019fairness, wachter2020fairness}, of which fairness mitigation techniques and toolkits can be an important part of, but are not ends in themselves. {As existing work has demonstrated, selecting an appropriate fairness metric and mitigation method depending on the context of the use case is challenging even for experienced AI practitioners \cite{lee2021landscape}.}
The AutoML paradigm of ML development will often (almost by definition) be unable to bring into account all the requisite factors, perspectives and constraints (which often require some domain expertise) that would otherwise need to be reconciled in addressing issues of fairness. Today's AutoML systems tend to limit customers' interaction with the system to a few specific decision points and present only limited to no information on how these systems work in operation and the complex processes behind the ML models generation and selection \cite{wang2019atmseer, xin2021whither, weidele2020autoaiviz, lee2019human, wang2019human}. This `blackbox' nature of AutoML operation results in a situation where users will often be unable to understand \textit{how} and \textit{why} AutoML systems make the choices they make \cite{weidele2020autoaiviz}, thus hindering their ability to effectively reason about and mitigate potential biases embedded in their outputs. 

\subsection{AI APIs}\label{Concerns:APIs}

While AutoML services aim to automate the creation of \textit{custom} machine learning models based on users' data and needs, AI APIs provide access to readily-available \textit{pre-built} AI models that allow users to instantly integrate AI capabilities into their applications. This instant availability, however, generally comes at the cost of customisation. In the AI APIs setting, generally the user has no control over the dataset used to train the model or influence over the specifics of the model itself.\footnote{Note that there are exceptions, for example, services such as Amazon's Rekognition Custom Labels and Google's Vision AutoML sit at the intersection of AutoML and AI APIs allowing users to partially tailor the service to their needs. To ease discussion, here we focus on AI APIs that are offered ``as is'' and do not allow for further customisation, though similar issues may well be relevant, particularly depending on the degree to which customisation is possible.} And while providers market these services with broad statements such as ``infuse powerful AI capabilities into your apps and workflows'' \cite{IBMWatsonAI}, ``get quality and accuracy from continuously-learning APIs'' \cite{AmazonAIServices} it is generally unclear what the performance of these services exactly is, how have they been trained, and how have they been evaluated. Below, we examine how the `one-size-fits-all' ideal that underpins AI APIs, coupled with the inherent opacity of these services,  can lead to potential bias and fairness issues and tensions when they are applied in the user-specific application contexts.

\subsubsection{Universality}
\label{Concerns:APIs:Representation}

It is well acknowledged that machine learning models, such as those offered by the API providers, are prone to exhibiting various kinds of discriminatory behaviours, including different error rates across demographic groups \cite{buolamwini2018gender, raji2019actionable, raji2020saving}, stereotyping \cite{zhao2017men, hendricks2018women} or minority groups under-representation \cite{shankar2017no, scheuerman2020we}. Many of these works conclude that these issues are primarily a reflection of training set deficiencies, namely the under-representation of certain parts of the input space \cite{buolamwini2018gender, bender2021dangers} and the lack of awareness, consideration and attention  of the people building these systems \cite{selbst2019fairness, madaio2020co, trewin2019considerations}. Consequently, a significant body of work argues for careful consideration of the social dimension in which the model will be applied, which therefore requires the use of 
training data that is appropriate for capturing the full diversity of the context \cite{raji2020closing, cobbe2021reviewable, suresh2019framework}. As we argue next, however, the nature of the AI API service provision model---which aims to provide  \textit{universal} `AI building blocks' that can be used and deployed in a range of customer applications, without knowing the particulars of the customer's specific usage contexts---may render these services inherently incapable of addressing the representation issue.

\begin{table*}
\parbox{.45\textwidth}{

\centering
\resizebox{0.4\textwidth}{!}{
\small
\begin{tabular*}{0.4\textwidth}{l|@{\extracolsep{\fill}}cccc} 
\toprule
\textbf{Company}   & \textbf{Asian}   & \textbf{Black}   & \textbf{Indian}  & \textbf{White}    \\ 
\hline
Amazon           & 8.39           & 9.21           & 7.85            & 7.20        \\
Baidu            & 5.34           & 7.89           & 6.90            & 7.54        \\
Face++           & 6.72           & 8.74           & 8.25            & 8.62            \\
Microsoft        & 5.26           & 5.88           & 5.76            & 5.19            \\
\bottomrule
\end{tabular*}
} \caption{Age estimation performance across race groups as measured by Mean Absolute Error (MAE) metric.}
}
\hfill
\parbox{.5\textwidth}{
\centering
\resizebox{0.47\textwidth}{!}{
\begin{tabular*}{0.55\textwidth}{l|cccccccc} 
\toprule
\multirow{2}{*}{\textbf{Company}} & \multicolumn{2}{c}{\textbf{Asian}} & \multicolumn{2}{c}{\textbf{Black}} & \multicolumn{2}{c}{\textbf{Indian}} & \multicolumn{2}{c}{\textbf{White}}  \\
                                  & \textbf{u. 60} & \textbf{60+}      & \textbf{u. 60} & \textbf{60+}      & \textbf{u. 60} & \textbf{60+}       & \textbf{u. 60} & \textbf{60+}       \\ 
\hline
Amazon                            & 4.87           & 19.98             & 6.31           & 18.5              & 4.7            & 15.66              & 5.34           & 16.75              \\
Baidu                             & 3.69           & 10.78             & 5.44           & 15.72             & 4.58           & 14.75              & 5.46           & 14.94              \\
Face++                            & 6.07           & 8.82              & 8.52           & 9.44              & 8.2            & 8.43               & 9.08           & 6.99               \\
Microsoft                         & 3.44           & 11.22             & 4.33           & 11.17             & 4.03           & 9.15               & 4.19           & 11.21              \\
\bottomrule
\end{tabular*}
}
 \caption{Age estimation performance across age and race groups as measured by Mean Absolute Error (MAE) metric.}
}
\end{table*}

\begin{table*}
\centering
\resizebox{\linewidth}{!}{\begin{tabular}{l|ccccc|ccccc|ccccc|ccccc} 
\hline
\multirow{2}{*}{\textbf{Company}} & \multicolumn{5}{c}{\textbf{Asian}}                                                       & \multicolumn{5}{c}{\textbf{Black}}                                                       & \multicolumn{5}{c}{\textbf{Indian}}                                                      & \multicolumn{5}{c}{\textbf{White}}                                                        \\
                                  & \textbf{\textless{}15} & \textbf{15-24} & \textbf{25-44} & \textbf{45-64} & \textbf{65+} & \textbf{\textless{}15} & \textbf{15-24} & \textbf{25-44} & \textbf{45-64} & \textbf{65+} & \textbf{\textless{}15} & \textbf{15-24} & \textbf{25-44} & \textbf{45-64} & \textbf{65+} & \textbf{\textless{}15} & \textbf{15-24} & \textbf{25-44} & \textbf{45-64} & \textbf{65+}  \\ 
\hline
Amazon                            & 1.38                   & 3.94           & 5.08           & 9.67           & 21.88        & 5.34                   & 5.25           & 5.8            & 9.85           & 19.85        & 2.88                   & 4.63           & 4.28           & 7.73           & 16.46        & 3.08                   & 4.64           & 5.09           & 9.31           & 17.15         \\
Baidu                             & 1.24                   & 1.83           & 4.22           & 8.34           & 11.11        & 3.29                   & 4.9            & 4.95           & 10.07          & 16.23        & 2.46                   & 4.8            & 4.13           & 7.94           & 15.12        & 2.5                    & 6.06           & 4.63           & 9.46           & 15.06         \\
Face++                            & 3.19                   & 6.04           & 7.27           & 8.17           & 8.76         & 13.06                  & 5.48           & 6.77           & 8.66           & 9.81         & 11.37                  & 6.11           & 7.02           & 8.45           & 8.32         & 13.59                  & 6.86           & 8.12           & 7.58           & 6.92          \\
Microsoft                         & 1.1                    & 2.61           & 4.18           & 6.09           & 12.15        & 2.09                   & 3.73           & 4.71           & 7.39           & 11.8         & 1.31                   & 5.21           & 4.35           & 5.33           & 9.64         & 1.92                   & 4.65           & 4.39           & 6.31           & 11.42         \\
\hline
\end{tabular}
}
\caption{Age estimation performance across age and race groups as measured by Mean Absolute Error (MAE) metric.}
\vspace{-2mm}\end{table*} 
Consider facial analysis services as an example, an area that has had some prior attention on issues of bias \cite{buolamwini2018gender, raji2019actionable, grother2019face}. These services, generally available  `off-the-shelf' to anyone, aim to determine an individual’s facial characteristics including physical or demographic traits based on an image of their face. However, human faces are not a homogeneous  {group}~\cite{khan2021one}; the set of images used to train and evaluate the underlying model will have a significant influence over the model's behaviour and reported performance. Without knowing specifically how and where their customers will use their services, the AI API providers are at constant risk of failing to envision, let alone account for, all the various contexts their models' might encounter. Consequently, they risk their services failing to be representative of particular regions, cultures, groups or individuals. 

 To illustrate this difficulty in ensuring and evaluating a model's general representativeness, we examine the performance of the APIs from Amazon, Baidu, Face++, and Microsoft for age estimation tasks\footnote{Note there is much scepticism (which we share) about models for such tasks; we use age estimation as our exemplar as it is both a service offered by many AIaaS providers and represents an intuitive way to present the broader point.} across race groups, using a subset of UTKFace dataset \cite{zhifei2017cvpr}. First, we consider the APIs performance on unitary subgroups. Table 2 shows that all APIs perform relatively well, displaying no obvious differences between different race groups. However, following Buolamwini \& Gebru's ``Gender Shades'' study and others \cite{buolamwini2018gender,raji2019actionable, raji2020saving}) and introducing \textit{intersectional} considerations (across categories or characteristics)---which are said to provide a more complete picture of the biases that may exist in an AI system \cite{buolamwini2018gender, raji2020saving}---reveals the existence of undesirable discrepancies. As shown in Table 3, disaggregating these subgroups into populations of under and over 60 years old reveals that all classifiers perform worse on the age estimation task for the 60+ population. Notably, Amazon's API age estimates on the 60+ subgroup exhibit an average error of almost 20 years for the Asian subgroup and over at least 15.5 years for all remaining groups. Further, as seen in Table 4, disaggregating those subgroups even further into UN's standard age brackets \cite{united1982provisional}, makes this lack of representation of certain subgroups even more apparent. This is referred to as `representation bias' (see \cite{suresh2019framework,lee2021algorithmic}), or an unintended skewing of model outcomes due to the dataset's representativeness. The consequences are a ``quality of service harm''~\cite{crawford2016artificial}, or the disparities in how well a model works for different groups of people, resulting in different experiences.

While some aspects of bias can potentially be reduced to \textit{some} extent, in some situations, for example by training models on more diverse and representative datasets \cite{raji2019actionable}, this assumes the relevant contexts were uncovered and accounted for. With no easily discernible and limited finite set of `protected classes' to rely on, there is no generally accepted limit to the number of potential categories bias can be evaluated on. Here we focused on the legally protected characteristic of  age, but we might have equally well examined images of people with facial tattoos, brown eyes, glasses, headwear or other attributes, which could also reveal certain biases (see e.g. \cite{denton2019image, georgopoulos2021mitigating}).
In general, and the what the above shows is that understanding and mitigating bias, fairness and other concerns of a model requires an understanding of \textit{salient factors of the context in which it will be applied}.

From an AI API provider's perspective, who seeks to offer a generic `one-size-fits-all' service, this therefore raises  the general question of whether eliminating all types of bias and making the model representative of all groups is ever possible? Without knowing \textit{how} their services will be used, \textit{where, when} and by \textit{whom}, despite best intentions, they might be unable to foresee all the various contexts in which their services will be used and thus likely fail to address the context-specific needs of a user use-case.

\subsubsection{Provider perceptions}
\label{Concerns:APIs:OneSize}

It is clear that the models a provider builds will be informed by or otherwise reflect (to varying degrees, depending on the circumstances) their view of the world. Given that AI APIs generally take a `one-size-fits-all' approach, where such services are made widely and generally available, there will inevitably be mismatches between the perspectives and considerations of a provider and those users who seek to integrate these services to drive their applications

For example, Terrance et al.~\cite{de2019does} found that publicly available object-recognition services are less effective at recognising household items that are common in particular geographies and low-income communities. Suggested factors for this included differences in their appearance or the contexts in which they appear, and the use of `English' as a base-language to describe particular items.
This example represents a real-world case showing that globally-available AI services, reflecting the provider's own perspectives and assumptions, can under-perform when applied in particular user contexts (here around particular geographies and income-levels).

Further, we argue there are some AI API services are too contextual, subjective, or ill-defined to form a general model for use by a range of actors, or indeed, even to be modelled at all. Using an extreme  example to illustrate, we find that two prominent AIaaS providers (Baidu and Face++) now include a beauty estimation feature as part of their facial analysis offerings~\cite{FacePlusPlus, Baidu}, promoting it in cosmetics and matchmaking sectors, but the feature is available for general use.
Beauty, however, is an inherently context-dependent and highly-subjective concept, which naturally varies between individuals, cultures, and regions.
While beauty estimation services immediately raise red flags for a number of reasons, we mention them as they exist, are openly available commercially, and starkly illustrate how AIaaS providers can make certain assumptions or take certain positions regarding the world. 
That is,  here, the providers of the service have decided on what `beauty' looks like and how it should be scored -- which will be reflected downstream in all their service user's applications.
Interestingly, we see that Baidu's service outputs a single `beauty-score' for any image, while Face++ gives two outputs to represent beauty scores ``from both male's and female's perspectives'' \cite{FacePlusPlus}; which illustrates not only how a provider's perspective impacts the services they offer, but also that these can differ from that of others, including other providers.

While it is generally the case that providers' views are encapsulated in their models (to varying degrees) for all services, this is especially problematic for concepts which are highly-subjective, and for which there is no generally agreeable `ground-truth'. 
Naturally this is the case for beauty scoring, which is something that cannot be generally modelled, and in our view---not least given the potential ramifications of such a service---it would be wholly inappropriate to try.

The broader overarching point (inappropriate services aside) is that certain assumptions, values and ideals may hold in some  contexts and situations but not in others. There will be many situations where a provider will not have considered the particulars of the social, economic, environment or other context of use; where a  provider decides, judges or takes a position on particular concepts that are not or cannot be universally agreed upon; or where it is challenging, infeasible or inappropriate to reduce a range of distinct concepts, definitions and groups  into a single model, let alone for models intended to operate generally (i.e. as a service) for use by others in a range of different contexts. Aiming to provide `universal' functionality, which abstracts away or otherwise does not account for real-world differences will inevitably lead to problems.

\subsection{Fully-managed AI Services}\label{Concerns:FullyManaged}

Finally, in the Fully-managed AI Services model, the control over the AI system is delegated almost entirely to a third party, who provides access to  packaged solutions, typically entailing some particular workflow that incorporates one or more ML models. Some prominent examples of this service category include HireVue (algorithmic recruitment) \cite{HireVue}, Luminance (AI for the legal profession) \cite{Luminance}, 
and Lunit (AI for radiology) \cite{Lunit75:online}, all of which offer access to complete applications and systems that are driven by AI to varying extents. 
Similar to the AI APIs, the ML models underpinning fully managed services are often offered on an ``as-is'' basis where no further user-led model retraining or larger customisation is possible, and are thus vulnerable to many of the challenges we have described in the previous section. 
However, the fully-managed service model contrasts with the AI API approach which typically concerns the more direct interactions between the user and access to the model itself, which is here further removed. Moreover, fully-managed services tend to deal with more numerous and\slash or complex set of functionalities---targeting niche industry-specific needs, in which contextual-specifics may be of even greater importance---potentially amplifying existing or incurring new kinds of risks and tensions. We elaborate these points below.

\subsubsection{Representative of whom?}

We explored above how the ``one-size-fits-all'' ideal underpinning AI APIs can lead to potential issues and tensions when these services are applied in the context their providers have not accounted for, or when the service aims to aggregate multiple inherently heterogeneous groups into one, resulting in a model that fails to accurately represent certain groups. Here, we argue that the complex, specialised subject matter of the fully-managed services and the generally opaque-manner in which they operate potentially amplify these risks {in two key ways} and make the question of their representativeness even more pertinent.

First, is the issue of the number of social categories and other factors that might be salient in deciding when determining a service's fairness, particularly as fully-managed services tend to embed a  range of processes and  workflows.
One prominent area in which fully managed AI services are becoming increasingly common is healthcare -- this includes services for radiology \cite{QureAI, Lunit75:online, AIMarketRadiology}, dermatology \cite{SkinAnalytics}, mental health \cite{Headspace, FeelTherapeutics} among others. These services offer access to AI-enabled workflows and platforms, incorporating one or more pre-built ML models and potentially various other inputs, that allow users (customers) to predict certain health conditions \cite{Aidence, Lunit75:online, QureAI}, discover patterns and changes in individuals' health \cite{FeelTherapeutics, Headspace} or receive medical recommendations and advice \cite{AIMarketRadiology, Headspace}. Similar to the AI APIs, with the services already largely pre-built and allowing limited further user customisation, the providers of these services need to again make assumptions about who the target users of their services will be and what training data they should use to build their models. As we have shown, this is challenging even in the context of single AI APIs, such as the Facial Analysis services, where evaluations relate to a set of common physical attributes such as gender, skin type, and the intersections thereof. The challenges of representativeness are likely only be exacerbated in more complex domains, such as healthcare diagnostics, where there may be a virtually unlimited number of individual, group, and social factors that can be relevant, from lifestyle and eating habits, to weather and chemical exposure, and particularities regarding where one lives \cite{challen2019artificial, TheGeogr51, park2018methodologic}.

Secondly, and further complicating the matter is the issue of \textit{testability}. Whereas AI API services are often openly accessible to anyone, which allows potential users as well as researchers and regulators to examine and validate these services to \textit{some} extent (e.g. by validating their performance against different  benchmarks and types of data \cite{buolamwini2018gender, scheuerman2019computers}), the access to fully managed AI services is by and large strictly limited to those directly engaging with the firm, or through provider-managed `demos' which often come with limits or restrictions on functionality and use (and thus might effectively prevent interrogation). Moreover, whereas the API paradigm allows for direct interaction with the model itself, for fully-managed services often the workflows are often more complex, possibly involving several models and other processes (technical or otherwise) interacting with each other. This can make it harder to examine these services and the specific aspects of concern. In all, these characteristics limit the scope for potential external oversight as well as prospective user's ability to compare and validate the representativeness and suitability of these services for their particular application, exacerbating the risks of potential fairness conflicts and tensions. 

\subsubsection{Fair how?}

Data choices notwithstanding, it is also important to consider what does it mean for these systems to be \textit{fair}? As an example consider algorithmic recruitment services (e.g. \cite{HireVue,pymetrics}), which provide users (customers) with access to a suite of candidate assessments (questions, video interviews, games etc.) which are then algorithmically analysed by the provider and their outputs combined to score and rank candidates. These systems are concerned with allocating or withholding opportunities or resources; therefore, fairness is of key concern. But {`fair'} can mean different things in different contexts to different people \cite{mulligan2019thing}. As we discussed in \S\ref{Background:Fairness} scholars have proposed a number of competing ways to quantify fairness \cite{verma2018fairness, dwork2012fairness, liu2017calibrated, joseph2016fairness}, some of which are mathematically incompatible with each other \cite{chouldechova2017fair, kleinberg2016inherent, friedler2016possibility}. 
In the typical machine learning development process, the creator of the system selects which definition of fairness to apply based on own understanding of the specific contextual needs of a given use case. However, this poses a unique challenge to the provider of an AI service because there may be multiple contexts and use cases for which different ways of measuring fairness may be applicable (see \S\ref{Concerns:AutoML:Constraints}, \S\ref{Concerns:APIs:Representation}). It also poses a challenge to the service user, who may not have all the information to assess for fairness. The user may have access to general information about candidates scores and the overall ranking, for instance, but the full details of the selection process, fairness metrics and mitigation measures used, tend to be managed by the provider and hidden from direct user oversight and control.

This can lead to potential tensions, where the definition of fairness chosen by the service provider is incompatible with the notion of fairness held by the customer, or the legal requirements of the market they operate in. For instance, in the now well-known case of COMPAS recidivism risk assessment software, ProPublica's investigative journalists argued that the system is biased against black defendants based on its failing of the ``False Positive Rate Parity'' definition of fairness \cite{ProPublica}. In response, the creator of the system (Northpointe) argued that its system was in fact fair according to ``Positive Predictive Value Parity'' definition of fairness, and it is the measure of fairness that ProPublica selected that was flawed \cite{ProPublica}. 
Similar issues can arise in other fully managed AI services, where user needs and expectations may not align with, or indeed, be incompatible with the assumptions and approaches taken by the provider.

As it stands, without defined sets of standards or laws on which fairness metrics need to be met and with the customers of AI services having no say in the definition of fairness to be followed, there is a risk that service providers might thus opt to adopt fairness definitions according to their own concerns, priorities, and interests.  For instance, in their evaluation of commercial algorithmic employment assessment services \cite{raghavan2020mitigating}, Raghavan et al. found that all evaluated providers who made concrete claims about fairness and debiasing practices of their systems did so with reference to the U.S. Equal Employment Opportunity Commission (EEOC) Uniform Guidelines \textit{4/5ths rule} \cite{bobko2004four}. This is understandable considering that majority of these providers are located in the U.S. and might thus be legally obliged to adhere to this rule. However, the customer base of these services is global and subject to wide range of different jurisdictions which may or may not follow similar guidelines. For instance, as Sánchez-Monedero points out \cite{sanchez2020does}, in the EU and UK, where no equivalent of 4/5ths rule exists in either statute or case law and where concepts of direct and indirect discrimination raise different considerations around algorithmic fairness~\cite{adams2022}, the provider's choice of 4/5ths rule as a measure of disproving bias is no longer so obvious, as other measures might be more appropriate. 
Given the global nature of services, and the multitudes of separate, potentially conflicting national guidelines and regulations to follow, providers will most likely consider the legal requirements of their own or key customers jurisdictions, but are unlikely to consider the full range of similar, overlapping and conflicting notions of bias, fairness, and discrimination in philosophical, sociological, legal and cultural context of every potential customer \cite{selbst2019fairness}. Again, by taking away the control of such context-dependent aspects from the customer, this fully managed AI service model risks imposing providers' world views and values onto their clients, which could lead to potential unintended consequences and tensions.
 \section{Discussion}\label{Discussion}
As AIaaS grows in prominence, so too does the risk of the real-world harm of these services. Indeed, as we have argued, there exists an inherent tension between the purportedly generic, `one-size-fits-all' ideal underpinning AIaaS and the nuanced, context-dependent nature of fairness. In aiming to offer AI services generically to potentially millions of customers, AIaaS providers need to make certain assumptions or choices about where and how their services will be used and by whom. At the same time, the world is vast and full of diverse groups that practise a range of traditions, customs and activities and share a variety of views and beliefs. As we have shown in this work, inevitably there will be situations in which the `generic' assumptions underpinning AI services will fail to address or come into conflict with the specific, context-dependent needs of a specific user's  use case (which the AIaaS providers are unlikely to have considered for all their potential customers) and in turn may render these services and their end-users' systems biased. 

We thus argue for caution and consideration in the development and use of AI services in practice. The AIaaS model has the potential to make AI more widely accessible, making it easier, faster and cheaper to deliver state-of-the-art AI and its benefits to a much wider range of applications and groups than might otherwise be possible. However, without proper care, these seeming benefits of `turn-key' AI availability and virtually limitless scalability can quickly risk potentially exacerbating the many bias problems and the broader ethical concerns around AI that various research communities have repeatedly cautioned against. 

Our focus so far has been on the risks and shortcomings of the AIaaS paradigm to draw attention to the area. 
We next identify potential ways forward and explore some considerations representing opportunities for discussion and future research in this space. 

\subsection{Provider Considerations}\label{Discussion:Provider}

\subsubsection{Appropriateness of the AI service}\label{Discussion:Provider:Appropriateness}

As we have detailed and shown, there are many situations in which the `one-size-fits-all' ideal of AIaaS can come into conflict with the contextual needs of a user's application, leading to potential risks and harms. Therefore, we join the growing number of calls urging researchers and developers to shift to a mindset of careful planning and evaluation in order to determine whether an attempt to build an AI model---or in this case an AI service---is appropriate in the first place \cite{selbst2019fairness, bender2021dangers, raji2020closing, mitchell2019model}. While this concern is relevant for any ML model development, it is particularly pertinent in the AIaaS context where potential harms might achieve a far larger scale \cite{cobbe2021artificial, javadi2021monitoring}. Before building a new AI service, it is important to consider to what extent (if at all) is it possible for the service to accurately model the social and technical requirements of the various contexts in which it will be deployed. \textit{Who} might be the target audiences and \textit{how} might they use the service? \textit{When} and \textit{where} might the service possibly be used (or misused)? Is it feasible or even possible for the model to accurately capture the social environment and its various needs, or would the modelling required be so complex as to be computationally intractable? Or indeed, is the topic too subjective or ill-defined to be one that is suitable or appropriate for general modelling?

As an illustrative example, an AI service may be designed for identity verification for users signing up for new financial products. The service providers should consider the possibility that this could be adapted for other purposes---such as age prediction for adult content, or policing and surveillance---and ask themselves whether this is appropriate and acceptable. Possible mitigative actions on the vendor's part may include 1) disclosure on the original purpose of the model, 2) documentation on the data and model, such as its representativeness in age, gender, and nationality, and 3) contractual and license restrictions for certain usage, e.g. cannot be used for illegal activities, for police surveillance, or even for some undesirable (but legal) purposes. While this may not prevent malicious actors from misusing the tool, the exercise of considering the potential alternative uses of the AI service could bring to light the contextual limitations to enable more effective disclosure and guidance to the users.

Furthermore, we propose that stakeholders involved in the development of AI services think through the potentially negative and harmful consequences that might arise from their services being applied in the context they have unaccounted for. As we discussed in \S\ref{Background:Regulation} the EU's proposed AI Act provides for risk management processes for \textit{certain} ``high risk'' AI systems~\cite{EURLex5241}; we propose that similar processes should be required for \textit{all AI services} due to the inherent risks of mismatch between systems' development by providers and deployment by users. If at any point, such a list of assumptions, caveats and limitations becomes overwhelming, then it could be a reasonable indication that the problem the given service aims to address is too contextual to be abstracted into a `generic' AI service and consequently stop the development of such service. 

\subsubsection{Transparency}\label{Discussion:Provider:Transparency}

As a part of the internal and external accountability processes, transparency can assist in facilitating algorithmic fairness. Apart from the role played by industry-wide regulations and standards, transparency mechanisms \cite{birchall2014radical, sambasivan2021re, ehsan2021expanding} should be embraced and implemented by developers and organisations committed to the development of AI services. Transparency which supports broader accountability processes on training sets, models and internal processes behind these services, as well as in the wider-sense of acknowledgement of known-limitations, past failures, and lessons learnt \cite{ sambasivan2021re, cobbe2021reviewable} can help move AIaaS from the opaque, unknown and potentially harmful `black-boxes' of today, to a more understandable, accepting of their own limitations, building blocks for affordable and widely accessible AI systems of the future. 

 We note that just as it is recognised that transparency won't necessary solve algorithmic accountability issues~\cite{krollaa,Ananny_Crawford_2018, raji2020saving}, transparency will not by itself address all AIaaS concerns. However, by embracing transparency at every stage of the AIaaS development process, including being upfront about the design contexts considered and any known limitations of the service while marketing it to potential users, some of the risks we have cautioned against in this paper could be reduced. There appears a clear role for adopting and extending various documentation approaches, be they relating to
 data and models (such as datasheets~\cite{gebru2018datasheets}, model cards~\cite{mitchell2019model} and  factsheets~\cite{arnold2019factsheets}), their interactions and interconnections (e.g. decision provenance~\cite{singh2019decprov}), and broader, holistic socio-technical system reviews (e.g. reviewability~\cite{cobbe2021reviewable} and traceability~\cite{kroll2021traceability}).
 
 Similarly, from providing varying levels of control and service-details depending on the user's proficiency with ML (particularly within the AutoML context) \cite{xin2021whither}, to partnering with external auditors \cite{krafft2021action, deng2022understanding, raji2019actionable} to help uncover potential gaps and limitations of the service, there are many opportunities for (research on) increasing transparency and improving collaboration between developers, customers, and stakeholders of AIaaS models and services. Importantly, these mechanisms should connect to internal and external institutional and governance mechanisms for account-giving and redress to ensure that issues are properly identified and addressed (whether to compliance and assurance teams within providers, to regulators and other external oversight bodies, or to customers of AI services who wish to ensure that required standards are being met in the services they are using).
 
\subsubsection{Gatekeeping}\label{Discussion:Provider:Gatekeeping}

It is worth exploring the options for providers in taking a more active role in setting out the intended usage of their services and mitigating the risks stemming from the repurposing of their services in untested or untoward contexts. For example, they might wish to limit their services to users from particular regions, jurisdictions or industries, which they believe best fit the capabilities and known limitations of their service, thus avoiding some of the risks in dealing with unknown contexts. Further, moving beyond just the issue of fairness, AIaaS providers might wish to monitor and restrict the usage of their services in order to mitigate the reputational risks and issues that could arise from the misuse of their services in driving controversial, inappropriate, or even illegal applications -- as Javadi et al. discuss \cite{javadi2021monitoring,javadi2020monitoring}.

Conversely, one could question whether it is sensible for AIaaS providers to decide on what constitutes an appropriate, ethical or untoward use case, or who should or should not have access to certain set of services. Though here we do not explore the various business, legal and other perspectives that go behind this concept of gatekeeping, we highlight this as an area requiring further consideration.

\subsection{Usage Considerations}\label{Discussion:Usage}
\subsubsection{Responsible AIaaS procurement and usage}

In \S\ref{Discussion:Provider:Appropriateness} we argued that before building a new AI service, AIaaS providers should first consider the potential risks and limitations to determine \textit{whether} and \textit{if} such developments should be undertaken at all. Similarly, we argue that the users of AIaaS services should also  carefully consider the appropriateness and potential risks of procuring AIaaS solutions to drive their applications. Notably, it is the users of AIaaS that have the advantage of being placed to know the {precise context} in which the AI service will be used, something the AIaaS providers are currently only generally able to {try} to anticipate. It is therefore worth exploring the particular ways in which AIaaS users could leverage this knowledge to their advantage, so to support and effectively empower them in assessing the suitability of an AI service in the context of their applications.

Central to this will be the service's transparency mechanisms which we have discussed in the previous section, whereby information is made available to service users. This could include transparency on a technical level such as the training data used, fairness metrics and measures applied, performance achieved or methods adopted, and the interactions between models; but also transparency in a more general sense such as details of their development practices, intended usage and known limitations
\cite{raghavan2020mitigating}. However, for this to work, it is crucial that the transparency provided is meaningful and effective for users, and thus raises opportunities for the HCI community to play an important role in realising these.

Additionally, for some AI services---particularly those of an AutoML or AI API type---there might be scope for potential customers to take on a more proactive role and examine the suitability of a given service through their own algorithmic investigations. As we have shown in this paper, it may be difficult (if not impossible) for AIaaS providers to exhaustively test and validate their services as free from all kinds of bias in all kinds of contexts \cite{raji2020saving, selbst2019fairness}. However, by conducting their own experimentation---not unlike the experiments we presented in \S\ref{Concerns:AutoML} and \S\ref{Concerns:APIs}---as part of their provider\slash service vetting process, the users of AI services could themselves be well-placed to validate service performance \textit{in context of their specific use case}. Indeed, a recent line of HCI research has explored the idea of ``{everyday
algorithm audits''} whereby users detect, understand, and interrogate problematic machine behaviours via their day-to-day interactions with algorithmic systems \cite{shen2021everyday, deng2022understanding, devos2022toward}. We argue that there will be benefits in including AIaaS in those discussions; for example, by exploring designs and methods that  better support users in conducting more effective service audits for bias, and means for meaningfully presenting and communicating such findings back to service providers. 
In practice, this will require understanding service users' needs as it regards fairness testing, providing effective interfaces and usable tools for such, and so on.

\subsubsection{Understanding user needs and challenges around AIaaS usage}
Despite growth in the development and dissemination of AIaaS services, there has been little research investigating how AIaaS customers \textit{actually} use these services in practice, nor on practitioners’ experiences and desires around AIaaS and the gaps between the capabilities of existing AI services and the needs of their users. Similarly, little is known about who the typical AIaaS user is, what level of ML-proficiency do they possess in practice, or in which markets they operate in. 
This limits the accessibility and usability of the AI service and presents a risk that these services and any supporting tools might be designed in a way that fails to align with user needs. 

These issues have been shown in more general ML-contexts; for example, studies exploring user (ML practitioner) attitudes and expectations regarding fairness toolkits showed a misalignment of tools with user expectations and needs, and generally being difficult to use (among other concerns), which not only makes such tooling less-effective, but can potentially lead to increased bias risks \cite{lee2021landscape, deng2022exploring}. Similarly, in one of the few studies exploring practitioner experiences using AutoML {
tools and platforms}, Xin et al. demonstrated a dissonance between those building AutoML tooling 
seeking to achieve \textit{``full automation''}  
and that of actual users for whom ``\textit{a complete automation is neither a requirement nor a desired outcome}'' \cite{xin2021whither}. Instead, they found that users expressed interest in a more {``human-in-the-loop''} AutoML---that aims to empower users in undertaking ML development, rather than replace them completely.

As such, we argue that the AIaaS space would benefit greatly from the attention of the HCI community, by studying how practitioners actually attempt to use these services in the context of real-world tasks. This could also include exploring their prior knowledge and understanding of ML processes, awareness of legal obligations, attitudes towards fairness, amongst other concerns. Such understandings might result in highlighting new opportunities for tooling, or devising a common vocabulary and effective modes of presentation for the providers and users of AIaaS to discuss and compare individual services, communicate issues, and assist users in identifying provider services that best align with the views and needs of their specific application contexts.
In general terms, there seems much scope for users to be better engaged as part of the design of AI services, which ultimately can help better support the proper use of AIaaS and make the benefits of AI more widely-accessible, while reducing the potential risks. On the other hand, poorly implemented or poorly explained AI services risk that a customer would select a suboptimal or inappropriate tool for their use case, or simply decide against adopting AIaaS at all. In all, there are clear opportunities for research in this space.

\subsection{Policy Directions}
\subsubsection{Roles and responsibilities}

As we have discussed, there are increasing moves towards regulating AI systems, including the EU's proposed AI Act which will apply to ``providers'' of AI services. As we note in \S\ref{Background:Regulation}, however, AIaaS challenges established understandings of the roles and responsibilities of providers and users; in some cases, the users (customers) of those services may themselves become considered as the ``provider'' under the Regulation for the underlying AI system in relation to their use of the service. As such, the compliance obligations (which include those relating to  bias and fairness) applying to providers will in those cases also apply to these ``user-providers'' (Article 28, EU AI Act). Yet such user-providers are downstream of the problems that the AI Regulation~\cite{eu2021proposal} seeks to address. As this paper makes clear, in a service context such-user providers will therefore lack anything like the information needed to meet their obligations; may not be in a position to understand or deal with such concerns, despite provider claims of ``\textit{no expertise required}''; and nor will they have the technical influence over such systems to be able to do so. 

Moreover, as we argue in this paper, general purpose models underpinning AIaaS can often exhibit structural issues that cannot be simply tweaked out of existence downstream by user-providers. Any biases and other undesirable characteristics ingrained in these services (by the AIaaS provider) will thus inevitably propagate downstream and there may be little downstream user-providers can do to address this; which is particularly problematic where---as this work makes clear---even systems {that} seem to be generically `fair' under some (generally broad, sweeping) measure,
may in fact exhibit bias issues when deployed in a specific context. As it stands, the downstream user-providers are given no help in scrutinising or holding to account upstream providers and instead are tasked with impossible compliance tasks. We therefore argue that the role of AIaaS providers and customers is an important issue warranting consideration in regulation targeting issues of AI fairness. 

\subsubsection{AIaaS regulation and guidance}
The emergence of the AIaaS model and its growing popularity calls for its explicit inclusion in ongoing debates on AI regulation, potentially necessitating the development of AIaaS-specific regulation, standards, and guidance. Indeed, given current regulatory directions \cite{eu2021proposal, FTC, jobin2019global}, there appears both scope and appetite for oversight bodies to set directions regarding various issues relating to AIaaS. {T}he consolidation of AI services around a few providers and the potential future widespread use of those services---although bringing challenges, as we have argued---may also mean that actually targeting {specific regulation, oversight, and enforcement mechanisms at certain} AI-related problems becomes easier than if many companies were developing their own AI systems in house.

For instance, regulations could set limits on acceptable types and uses of AI services and around monitoring for misuse of AI services for illegal or potentially harmful purposes, a recognised concern~\cite{javadi2020monitoring, javadi2021monitoring}. Such regulations could specify the role for various relevant actors---such as regulators, industry, and civil society---in complying with and overseeing these limits. In doing so, regulation may require explicit bans on types or uses of AI services which are deemed to pose a high risk to the rights, freedoms, and interests of people potentially affected by them, or more lenient sets of requirements, standards, and best practices for other, less risky systems. These could relate to, for example, the production and suitability of training data, to training and testing procedures and other areas of systems' development, 
as well as other aspects of a service offering with a view to identifying and addressing biases, in a manner that properly accounts for the position of the user as a customer of such services.

\subsubsection{Facilitating regulatory oversight}
As we have argued, transparency and accountability regimes are also needed to provide the information, mechanisms, and processes that allow the design and functioning of AI services to be better understood and to facilitate effective challenge and oversight by regulators, users (i.e. customers), and others. Mechanisms that enable regulators to obtain further information where needed would also help, as would mechanisms enabling users and regulators to hold providers to account and take corrective action where needed. While there is a growing interest in auditing algorithmic systems \cite{raji2019actionable, raji2020closing, raghavan2020mitigating, wilson2021building, shen2021everyday}, this field generally lacks defined standards and records could be incomplete or not relating to information needed for accountability purposes \cite{raji2020saving, bender2021dangers}. There may therefore be a role for regulators in defining standards and ensuring consistency in auditing practices so as to assist with facilitating oversight, particularly those cognisant of the `supply chain' that AIaaS inherently entails, and the nature of the relationships of the actors within that. Indeed, such activities also presents an opportunity for the HCI community, leveraging and extending work on how to provide means and tooling for meaningful transparency to support accountability~\cite{cobbe2021reviewable, norval2022disclosure, ehsan2021expanding, heger2022understanding, anik2021data}.

While we concretely highlight issues and provide suggestions to indicate the some potential avenues forward, the specifics regarding external oversight and scrutiny over the use, reliance, monitoring and actions of AI services require further consideration. Future research of an academic and policy nature may explore these issues in greater depth so as to develop more proposals for intervention in this space.
 \section{Conclusion}

In this paper we explored and elaborated a range of potential fairness concerns posed by AIaaS. Specifically, we introduced, reviewed and systematised the AI services landscape, proposing a general taxonomy of service type characteristics from a service user perspective. Further, we have examined each of the different types of AIaaS services using a combination of experimental evaluation and illustrative examples, outlining the bias and fairness concerns and practical challenges they bring to the fair ML discussion. We have highlighted the broader implications that the emergence of AIaaS has for issues of fairness and AI governance, suggesting potential research opportunities and ways forward.

The adoption of AIaaS will likely only continue to grow. As such, it is critical to address the open challenges, limitations and concerns this service paradigm presents. By providing this initial exploration of AIaaS fairness considerations, we hope to raise awareness of the variety of challenges in this space, and foster a discussion on what can be done to mitigate thee risks.
 
\begin{acks}
We acknowledge the financial support of the UKRI Engineering and Physical Sciences Research Council (EPSRC) (EP/P024394/1 and  EP/R033501/1), Aviva, and Microsoft through the Microsoft Cloud Computing Research Centre.
\end{acks}

\bibliographystyle{ACM-Reference-Format}
\bibliography{sample-base}

\end{document}